\documentclass{article}

\usepackage{microtype}
\usepackage{graphicx}
\usepackage{hyperref}

\usepackage{icml2026_arxiv}

\usepackage{amsmath}
\usepackage{amssymb}
\usepackage{xcolor}
\usepackage{array}
\usepackage{placeins}

\usepackage[capitalize,noabbrev]{cleveref}

\def\BB{{\cal B}}

\def\DD{{\cal D}}

\def\SS{{\cal S}}

\def\IR{{\mathbb R}}

\def\IF{{\mathbb F}}

\newcommand{\rmsbar}[2]{\textcolor{#1}{\rule{#2}{4.5pt}}}

\icmltitlerunning{Neural Operators as Efficient Function Interpolators}

\begin{document}


\twocolumn[
    \begin{flushright}
        CERN-TH-2026-096 \quad
        CCTP-2026-8 \quad
        ITCP-2026-8
    \end{flushright}
    \vspace{0.2em}
  
  \icmltitle{Neural Operators as Efficient Function Interpolators}

  \icmlsetsymbol{equal}{*}

  \begin{icmlauthorlist}
    \icmlauthor{Vasilis Niarchos}{crete}
    \icmlauthor{Angelos Sirbu}{crete}
    \icmlauthor{Sokratis Trifinopoulos}{cern,uzh}    
  \end{icmlauthorlist}

  \icmlaffiliation{cern}{Theoretical Physics Department, CERN, Geneva, Switzerland}
  \icmlaffiliation{uzh}{Physik-Institut, Universit\"at Z\"urich, 8057 Z\"urich, Switzerland}
  \icmlaffiliation{crete}{Department of Physics,  CCTP and ITCP, University of Crete, 71303, Greece}
    

    \icmlcorrespondingauthor{Angelos Sirbu}{asirbu@physics.uoc.gr}


  \icmlkeywords{Neural Operators, Function Approximation, Interpolation, Nuclear Physics, Binding Energy}

  \vskip 0.3in

]
\printAffiliationsAndNotice{~}

\begin{abstract}
Neural operators (NOs) are designed to learn maps between infinite-dimensional function spaces. We propose a novel reframing of their use. By introducing an auxiliary base-space, any finite-dimensional function can be viewed as an operator acting by composition on functions of the base-space. Through a range of benchmarks on analytic functions of increasing complexity and dimensionality, we demonstrate that NOs can match or outperform standard multilayer perceptrons and Kolmogorov--Arnold Networks  in accuracy  while requiring significantly fewer parameters and training time. As a real-world application, we apply a two-dimensional Tensorized Fourier Neural Operator (TFNO) to the nuclear chart, learning a correction to state-of-the-art nuclear mass models as a partially observed residual field. A TFNO ensemble reaches a held-out root-mean-square error of $198.2\,$keV, placing it among the best recent neural-network approaches while retaining high parameter efficiency and short training times. More broadly, these results introduce NOs as a scalable framework for finite-dimensional function interpolation, from analytic benchmarks to structured scientific data.
\end{abstract}

\section{Introduction}
\label{sec:intro}

Interpolating unknown functions from sparse evaluations is a fundamental task across science and engineering. Classical methods---linear, polynomial, and spline interpolation---are well understood but often struggle with high-dimensional or highly oscillatory targets. Neural networks (NNs), established as universal function approximators~\citep{Cybenko:1989iql,Hornik:1991sec}, have become a popular alternative, though they can depend sensitively on the discretization of their training data and can be prone to overfitting. Other architectures, like the Kolmogorov--Arnold Networks (KANs)~\citep{liu2025kan}, inspired by the Kolmogorov--Arnold representation theorem, offer an alternative with better interpretability, but can be computationally expensive to train. Recently, connections between interpolation theory and neural networks were also explored in~\citep{INN2025,guo2025interpolating}. 

Neural operators (NOs)~\citep{li2021fourier,Lu2021DeepONet,kovachki2023neuraloperator} represent a fundamentally different paradigm: they learn maps between \emph{function spaces} rather than maps between finite-dimensional vectors. Designed primarily for solving parametric partial differential equations (PDEs), NOs---like the Fourier Neural Operator (FNO)~\citep{li2021fourier} and its tensorized variant (TFNO)~\citep{kossaifi2024multigrid}---possess the remarkable property of \emph{discretization invariance}. Once trained on coarse data, they can be evaluated at arbitrary resolution without retraining (zero-shot super-resolution). Related operator architectures include DeepONets~\citep{Lu2021DeepONet}. 

In this work, we ask: \emph{Can Neural Operators be repurposed to address the simpler but ubiquitous task of function approximation/interpolation?} 
A similar question was already considered in Ref.~\cite{huang2025operator}. In that paper, NOs were modified to perform the additional vector-to-vector, vector-to-function and function-to-vector tasks. A vector-to-vector task, which is simply a function, is also the main focus of the present work. A vector-to-function task is a parametric family of functions and a function-to-vector task is a functional. We will not consider the last two cases here. Compared to \cite{huang2025operator}, our approach is different because we do not alter the NO architecture; we simply reinterpret the training and inference process to learn functions in a `non-local' way across samples of higher-dimensional subspaces, instead of learning them point-wise with samples of individual points as standard MLPs do.

As a concrete real-world benchmark, we apply the operator-as-interpolator viewpoint to nuclear-mass prediction. The task is especially consequential because nuclear masses are not only fundamental to nuclear stability but also enter directly into a wide range of calculations across physics, including astrophysics and particle physics. Rather than repeating the full physics motivation here, we treat the task as completion of a partially observed residual field on the $(Z,N)$ chart relative to state-of-the-art nuclear models, in particular Weizs\"acker--Skyrme version-4 (WS4)~\citep{Wang:2014qqa}. Evaluation is performed strictly out of sample, which is essential in nuclear-mass learning because even small amounts of target leakage can make quoted root-mean-square (RMS) values artificially optimistic. This provides a demanding test of whether NOs can exploit nontrivial structure in scientific data beyond controlled analytic benchmarks.

The precise setup is introduced in Sec.~\ref{sec:method}. Our main contributions are:
\begin{enumerate}
    \item A reformulation of NOs that recasts finite-dimensional function approximation as operator learning (\cref{sec:method}).
    \item Multiple benchmarks showing NOs are accurate and parameter- and training-efficient interpolators, competitive with MLPs and KANs on functions of varying complexity and dimensionality (\cref{sec:benchmarks}).
\item An application to nuclear physics, where a $\sim 150\,\mathrm{k}$-parameter 2D TFNO learns corrections to the WS4 mass model. Under pooled five-fold out-of-fold (OOF) evaluation, an ensemble trained embarrassingly in parallel reaches $198.2\,\mathrm{keV}$, making TFNOs competitive with the best recent leakage-free single-task methods in the literature. In addition, we find that this performance is achieved while the model remains lightweight to train (\cref{sec:nuclear}).

\end{enumerate}

\section{Method: Functions as Operators}
\label{sec:method}

\paragraph{An auxiliary base-space construction.}
Consider a target function $f:\DD_{\rm in} \to \IR^{d_{\rm out}}$ with domain of definition $\DD_{\rm in} \subset \IR^{d_{\rm in}}$. We introduce an auxiliary base-space $\cal{B}$ (with dimension $d_\BB$) and functions $x:{\cal B} \to \DD_{\rm in}$ in a suitable auxiliary Banach space $\SS_{\rm aux}$. We define an operator 
\begin{equation}
\label{aa}
\IF:\{ x: {\cal B} \to \DD_{\rm in} \} \to 
\{ f\circ x : \BB \to \IR^{d_{\rm out}} \}
\end{equation}
that acts on functions of $\SS_{\rm aux}$ by composition with the target function $f$,
\begin{equation}
\label{ab}
\IF[x](s) = f(x(s))~, ~~ s\in \BB
~,
\end{equation}
to produce functions from the auxiliary base-space $\BB$ to $\IR^{d_{\rm out}}$. Our main premise is that by learning the operator $\IF$, we are effectively learning the original target function $f$. Instead of approximating $f$ directly, we approximate the operator $\IF$ and we choose to do so using a NO.

\begin{figure*}[t!]
    \centering
    \includegraphics[width=\linewidth]{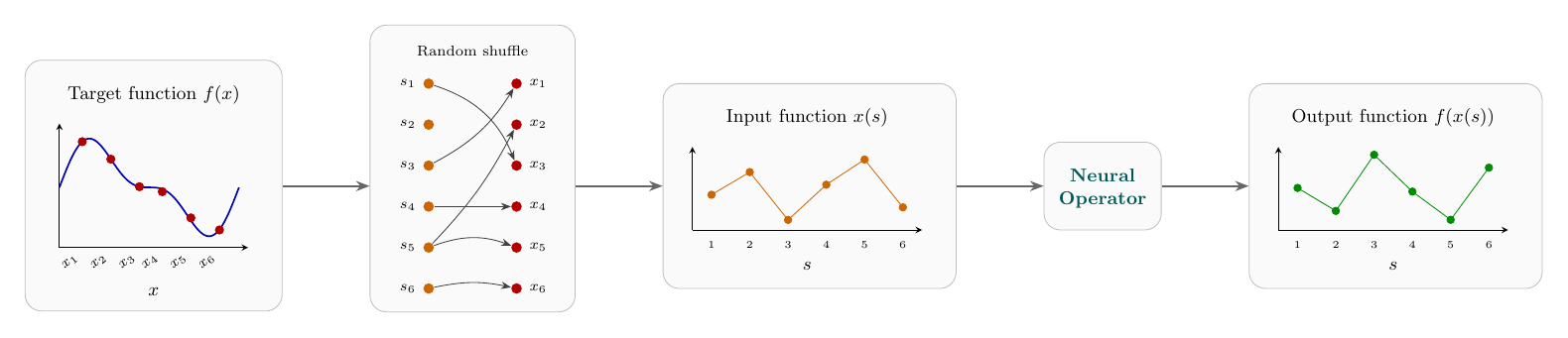}
    \caption{Diagrammatic representation of the NO pipeline for a one-dimensional base space.}
    \label{fig: pipeline diagram}
\end{figure*}

Practically, when we train this NO we proceed in the following manner. Assume we are given a set of training data $\{(x_i, f(x_i))\}_{i=1}^{N}$ for the target function $f$ and that we discretise $\BB$ on a grid with $r$ points. We use the training data to generate discretised input functions $x(s)$ by assigning groups of $r$ values of $x_i$ (and $f(x_i)$) to $r$ discrete values of $s$. In Fig.~\ref{fig: pipeline diagram} we provide a diagramatic respresentation of this process. This partition is repeated to create different samples (with potential overlaps) that cover the full training dataset. Typically, we implement this scheme with a random selection process.

\paragraph{Training and inference choices.}
Clearly, this construction involves several elements that can be chosen in different ways. 

One of them is the choice of the base-space $\BB$. By choosing its dimension $d_\BB$ so that $0\leq d_\BB \leq d_{\rm in}$, a scan over different (one-to-one) functions $g$ allows us to learn $f$ across different $d_\BB$-dimensional subspaces. Hence, different choices of $\BB$ (with potentially different discretization) amount to different types of learning reflecting different training strategies for the NO. For example, we can choose to learn $f$ using input functions $x(s)$ sampling each time a different local compact subset of $\DD_{\rm in}$ with minor overlaps.  Alternatively, we can use input functions $x(s)$ sampling (non-compact) subsets with repeated overlaps. We have observed that noisy implementations of the latter approach allow the operator to learn more efficiently the
global structure of the target function.

Another choice has to do with the specifics of the inference method. For a trained MLP that models a function, a prediction amounts to a simple evaluation of the MLP on a point of interest. For a trained NO both the input and the output are functions. As a result, on a $\BB$-grid with $r$ points, the output contains $r$ evaluations of the function of interest. Does the quality of the prediction depend on the type of input or the size of the discretization? We have observed that inference with input functions similar to those constructed during training (e.g.\ noisy functions both in training and inference) generically produces reliable predictions with low dependence on the number of discretization points. The latter fits nicely with the discretization-invariant properties of the NOs.  

\paragraph{Three implementation examples.}

{\it Zero-dimensional base.}
A limiting example of the construction arises when the base-space $\BB$ consists of a single point $s_0$. In that case, the functions $x(s), f(x(s))$ are a single set of vectors effectively collapsing the NO architecture to that of an MLP. For example, for an FNO, the spectral convolution in each layer--- which acts by pointwise multiplication in Fourier space---reduces to a standard linear transformation. For a \emph{Tensorized} FNO (TFNO)~\citep{kossaifi2024multigrid}, this layer reduces to a \emph{tensorized} linear layer, where the weight matrix is represented as a low-rank tensor decomposition (e.g., Tucker or canonical polyadic). In the benchmarks of the next section, we will call this tensorized MLP a ``zero-dimensional NO'' (0D-NO). To the best of our knowledge, the comparative efficiency of MLPs augmented with tensorized layers in function approximations 
has not been studied systematically in the past, and we take this opportunity to include them in our analysis.

\paragraph*{\it One-dimensional base.}
Another example that will be considered in the benchmarks of the next section is the case where $d_\BB=1$. In this context, we are learning functions along one-dimensional curves.

\paragraph{\it Two-dimensional base.}
In the Nuclear binding energies application of section \ref{sec:nuclear}, we aim to learn a 2-dimensional function of the number of protons and neutrons in nuclei. In that case, we consider a two-dimensional base-space.

\section{Benchmarks on Analytic Functions}
\label{sec:benchmarks}

Our objective in this modest benchmark is to compare the performance of 0D-NOs, 1D-NOs, MLPs and KANs on several functions of different dimensionality and complexity. 
We are particularly interested in exploring the comparative ability of (T)FNOs to learn complicated (and potentially unstructured maps like noise, where MLPs typically struggle) and their capacity to interpolate in higher dimensions from limited data.
 
For this purpose, we consider the following functions:
\begin{enumerate}
    \item The real part of finite partial-wave expansions,
    \begin{equation*}
        f_L(\theta) = \frac{1}{2}\sum_{\ell=0}^L (2\ell + 1)\, \sin{2\delta_\ell}\, P_\ell(\cos{\theta})
    \end{equation*}
    in $0<\theta<\pi$. To be concrete, we focus on partial-wave expansions of three increasing orders $L=3, 7, 13$. To create synthetic data, we sample the phases $\delta_\ell$ from a uniform distribution within the interval $[0, 2\pi)$.

    \item The Heaviside function on the interval $[0,1]$ with a step discontinuity at $x=0.5$.

    \item A piecewise combination of different Gaussian curves and a step discontinuity,
    \begin{equation*}
        f(x) = 
        \begin{cases}
            0.8\, e^{-\frac{(x+4)^2}{0.2}} + 0.2\, e^{-\frac{(x+1)^2}{0.1}}, & -5 \leq x \leq 0.7 \\
            0.5, & 0.7 < x < 3.7 \\
            0.6\, e^{-\frac{(x-6)^2}{0.01}} - 0.4\, e^{-\frac{(x-10)^2}{0.3}}, & 3.7 \leq x \leq 15
        \end{cases}
    \end{equation*}
    which combines the features of the previous two examples.
    
    \item 1D discrete uniform noise, sampled from the interval $[-2,3]$, on a uniformly 100-point grid in $[0,1]$. 

    \item Hypergeometric functions, $_p F_q(a_1,..,a_p;b_1,...,b_q;z)$ as $(p+q+1)$-dimensional functions of all their parameters, with $1\leq a_i, b_i \leq 2$ and $0\leq z\leq 0.75$. For illustration, we focus on the 4D function $_2 F_1(a,b;c;z)$ and the 6D function $_3 F_2(a, b, c; d, e; z)$. 
\end{enumerate}

\paragraph{Hyperparameter setup of all models.}

The architecture of the models is the same in all examples and chosen in a generic manner with the single restriction that all models have a comparable number of parameters (see Tab.~\ref{tab:params}). Aiming to capture generic features of the performance of the models, we do not perform any task-adapted hyperparameter tuning.

On the NO side, we consider a relatively small Tensorized Fourier Neural Operator (TFNO), implemented in {\tt PyTorch} via the {\tt neuraloperator} library~\cite{kossaifi2025librarylearningneuraloperators}. Our models have two layers with 16 Fourier modes, 32 hidden channels, and 64 lifting and projection channels. To tensorize the spectral convolution layer, we use Tucker factorization, with rank $10^{-2}$. Both for the 0D- and 1D-TFNO we set the number of input channels to the dimensionality, $D$, of the target function. Since we are working only with scalar functions, we fix the number of output channels to one. For all hidden layers we use GELU activation functions. To improve performance, we normalize both the input and output training data, using standard scalers. 

For the MLP, we consider a simple sequential neural network consisting of 4 identical hidden layers with 53 nodes. For all hidden layers we use GELU activation functions. In the benchmark plots, the legend label ``NN'' refers to this MLP baseline.

Finally, for the KAN we use a model that contains three identical hidden layers with 19 nodes. In order to save time and resources, we implement the above KAN architecture in {\tt fastkan}~\cite{li2024kolmogorovarnoldnetworksradialbasis}, which is an improved version of the {\tt pykan} implementation of KANs~\cite{liu2025kan} using a radial spline basis to accelerate runtime during training. Due to that radial-spline implementation however, single-input KANs (that is, for 1D functions) cannot be implemented in {\tt fastkan}. Hence, for 1D functions we implement a $[1, 19, 19, 19, 1]$-KAN in {\tt pykan}. Similarly to the MLP architecture, we have chosen this architecture so that the KAN has a comparable total number of parameters to the other models.

\begin{table}[h!]
    \centering
    \begin{tabular}{ccccc}
          & 1D-TFNO & 0D-TFNO & MLP & KAN \\
         \hline
        params & 8917 & 8917 & 8746 & 9120
    \end{tabular}
    \caption{Total number of parameters for each 1D model.
    }
    \label{tab:params}
\end{table}

For all models, we use an {\tt AdamW} optimizer with learning rate $10^{-3}$ and weight decay $10^{-5}$, together with {\tt ReduceLRONPlateau} learning-rate scheduler. We train all the models for 1000 epochs, using the Root Mean Squared Error (RMSE) loss.  All benchmark computations were performed on an Intel Core i7 CPU.

\begin{figure*}[t!]
    \centering
    \includegraphics[width=\linewidth]{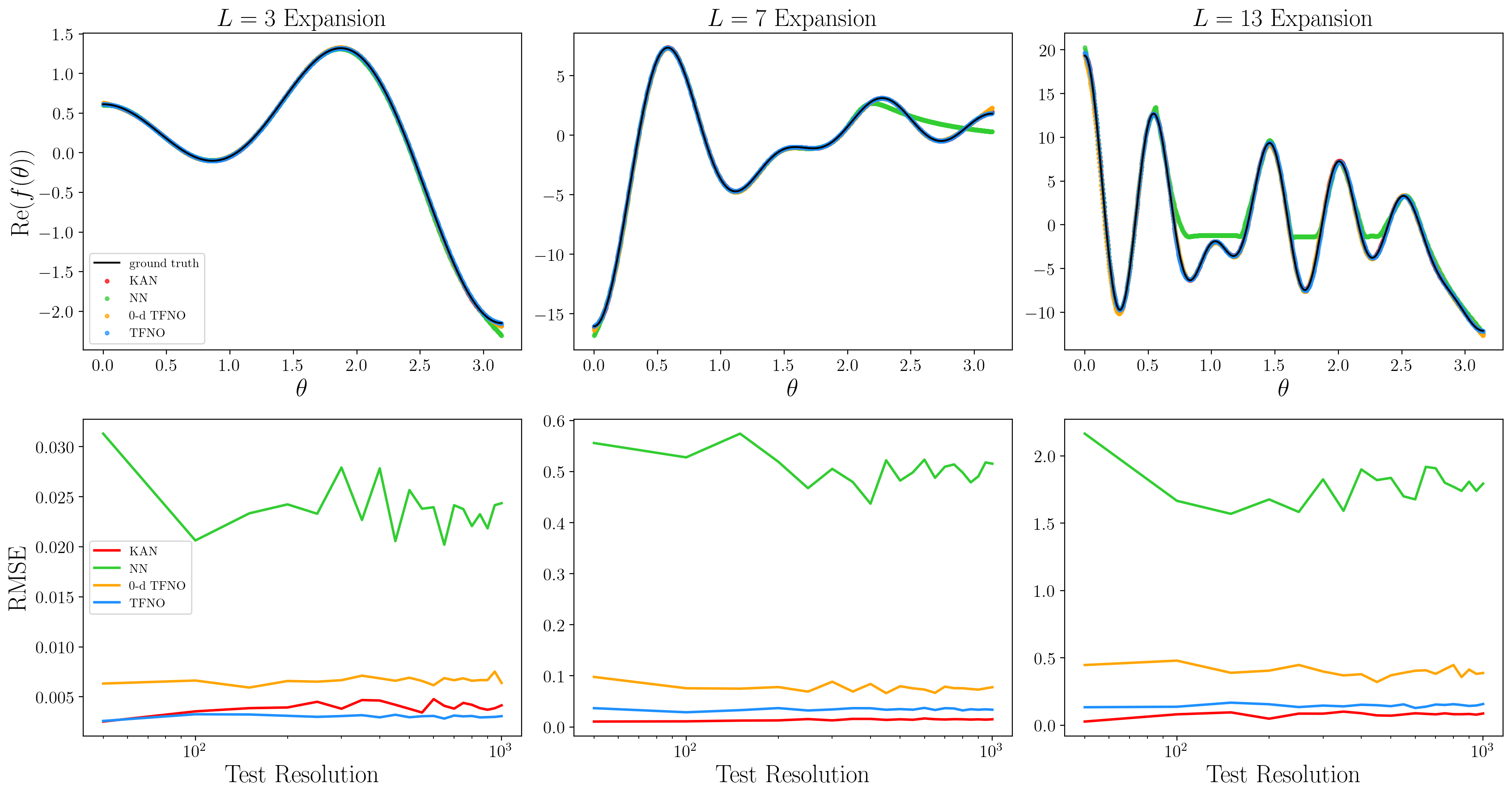}
    \caption{Benchmarks for order $L=3,7,13$ partial wave expansions. Top-row plots: Predicted maps of all models against the angle $\theta$. Bottom-row plots: test-set RMSE vs number of test points.
    }
    \label{fig: pw benchmarks}
\end{figure*}


\paragraph{Training data.}
To generate training data, we construct a grid of $N$ random points on the domain of definition of the target functions, sampled from a uniform distribution. We train the MLP and KAN on the raw training grid. To construct the training functions of the 1D-TFNO, we rearrange the points of the training grid into a stack of $r$-dimensional vectors, where $r$ is the grid resolution of the NO input functions. In the benchmarks of 1D functions, we typically use $N=100$. For the 1D-TFNOs the training on these 100 points is performed by constructing 2500 input functions with grid resolution $r=50$. We split the 1D-TFNO training set into batches of 64 samples. Since the sample size of the grid is small, we do not split it into batches in the training process of the 0D-TFNO, MLP and KAN.

\paragraph{Evaluation.}
To test the performance of each model we generate a series of random uniform grids with increasing numbers of points and evaluate the RMSE of each model, on each set. We then plot the test-set RMSEs against the number of points of the test sets, and compare the models against the RMSEs that they achieve and their stability across test set sizes. For the 1D-TFNOs, each point on the RMSE plot comes from the evaluation on a single input function with grid resolution equal to the number of evaluation points.

In addition, we plot the predictions, either against the function argument (in 1D examples), or against the ground truth (in higher-dimensional examples). 

\paragraph{Results.}
{\it Partial wave expansions.} 
The corresponding predictions of all the models are presented in Fig.~\ref{fig: pw benchmarks}. In all three examples, the ranking of the models is similar in terms of the recorded RMSE, with the KAN and the 1D-TFNO being the most competitive performers. In the bottom row of plots in Fig.~\ref{fig: pw benchmarks}, the KAN appears to outperform the 1D-TFNO in the higher angular momentum regime ($L=7,13$), although both models produce comparable RMSE. The 1D-TFNO interpolates smoothly and exhibits impressive stability of the RMSE across a range of different test sizes. Since all these point-evaluations are a single input-function evaluation for the 1D-TFNO (with different grid resolution each time), this feature is likely related to the zero-shot superresolution abilities of FNOs. We also notice that the 0D-TFNO outperforms the MLP.

{\it Heaviside.}
The results on the Heaviside function are presented in Figs.~\ref{fig: step benchmarks} and \ref{fig: heaviside rmses} in App.~\ref{app:plots}. In terms of the full RMSE curves, the performance of all models across the whole step function is comparable (see Fig.~\ref{fig: step benchmarks}). However, when we focus close to the discontinuity at $x=0.5$ we observe visible differences. Below $x=0.5$ (see the left plot in Fig.~\ref{fig: heaviside rmses}) the best performer is the 1D-TFNO and the worst one the MLP. Above $x=0.5$ (right plot in Fig.~\ref{fig: heaviside rmses}), the MLP performs better, but the 1D-TFNO remains competitive.

\begin{figure*}[t!]
    \centering
    \includegraphics[width=0.95\linewidth]{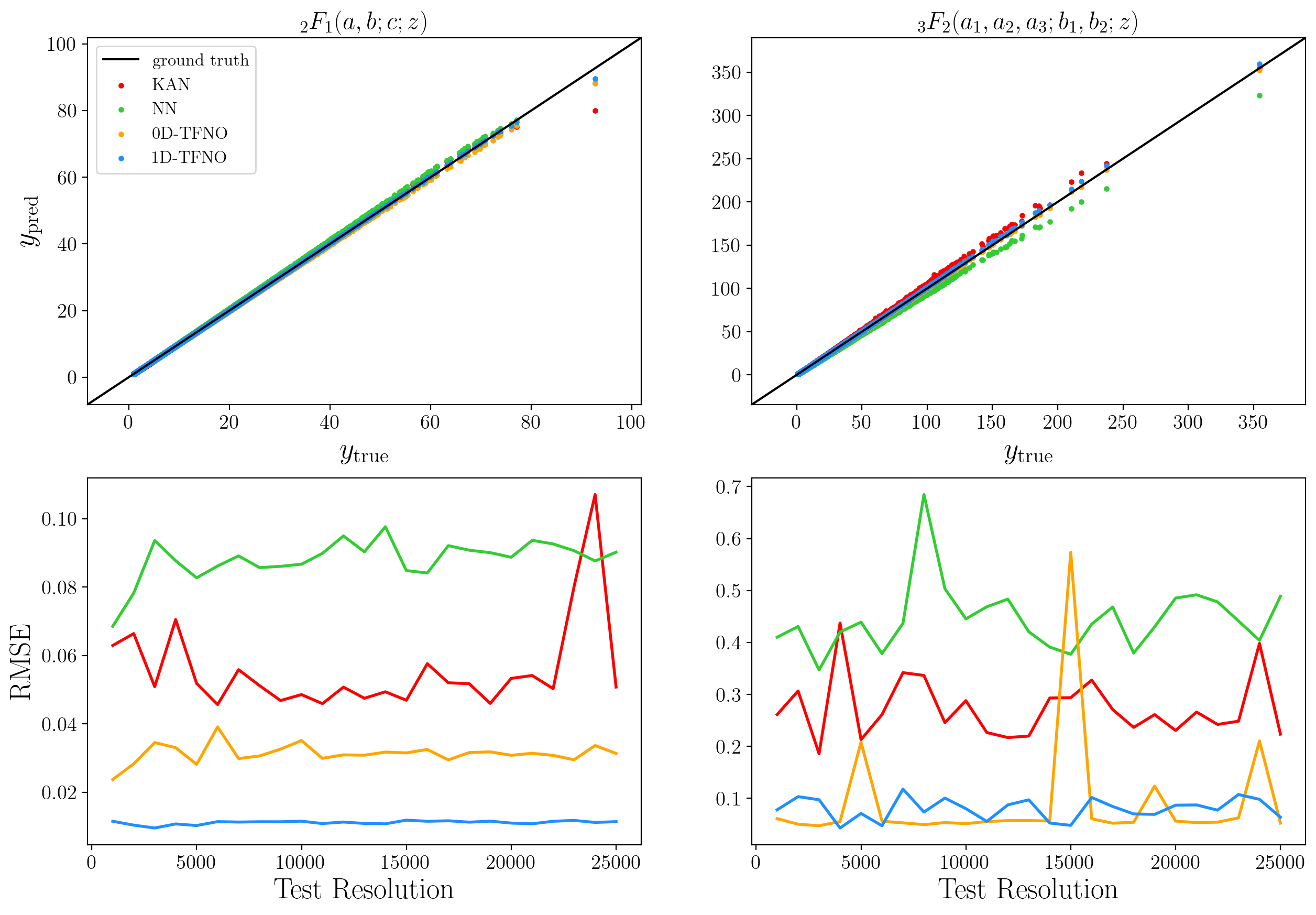}
    \caption{Benchmarks for hypergeometric functions. Top-row plots: predictions vs ground truth. Bottom-row plots: test-set RMSE vs number of test points. 
    }
    \label{fig: hyp geom}
\end{figure*}


{\it Piecewise combination.}
In a more complicated 1D function with a combination of multiple features we observe again that the 1D-TFNO is among the top performers. The 
results on the piecewise combination of Gaussians and step functions are displayed in Fig.~\ref{fig: piecew comb} in App.~\ref{app:plots}. In this case, all models were fitted on a random uniform grid of 500 points, which was split into batches of 128 samples. The MLP struggles with the high-frequency features and is overall comparable to the 0D-TFNO, but inferior to the 1D-TFNO and the KAN.

{\it 1D uniform noise.}
The 1D-TFNO is also exceptional in fitting the complicated, random structure of noise. The performance of all the models in 1D uniform noise is summarized in Fig.~\ref{fig: noise} in App.~\ref{app:plots}. The MLP produces a smooth, averaged map (as expected from spectral bias). The 0D-TFNO is similar, but appears to learn better some of the sharp features of the noisy function near the endpoints of its domain. The KAN captures some of the sharp features of the noisy function 
better than the MLP and the 0D-TFNO, but still produces large residuals in some regions.

{\it Hypergeometric functions.}
In the left two plots of Fig.~\ref{fig: hyp geom}, we present the benchmarks on the hypergeometric function $_2F_1(a,b;c;z)$ (viewed as a 4D function of the variables $(a,b,c,z)$). All models were trained on a random uniform grid of 75k points. For the 1D-TFNO, in particular, we constructed 4.5k input functions, with grid resolution $r=150$, and split the training set into batches of 128 samples. For the 0D-TFNO, the MLP and KAN, we split the training set into batches of 1536 samples.

In the right column of plots in Fig.~\ref{fig: hyp geom}, we present the corresponding results for the hypergeometric function $_3F_2(a,b,c;d,e;z)$ viewed as a 6D function. All models were trained on a random uniform grid with 150k samples in this case. Similar to the $_2F_1$ function, for the 1D-TFNO we constructed 4.5k input functions, with grid resolution $r=150$, and split them into batches of 128 samples. For the remaining models, we split the training set into batches of 3072 points. 

Our models rank similarly in terms of RMSE, for both functions, and once again the 1D-TFNO is consistently among the top performers. In the $_2F_1$ example, the 1D-TFNO exhibits the lowest, and impressively, the most stable RMSE. The second best performer, in terms of RMSE, is the 0D-TFNO followed by the KAN. The top-left plot of Fig.~\ref{fig: hyp geom} helps explain the fluctuations observed in the KAN's RMSE. By inspecting the predictions of the KAN against the ground truth, we notice that the KAN struggles to fit large values of $_2F_1$ (which correspond to the region near the edge corner $(a,b,c,z)=(2,2,1,0.75)$ of the function's domain) and tends to underestimate them. 

Likewise, the 1D-TFNO and 0D-TFNO rank first in the $_3F_2$ example with similar RMSE. Compared to the previous example, the RMSE of the 1D-TFNO exhibits slightly more fluctuations, but is still impressively stable and consistent across different test sizes. The RMSE of the 0D-TFNO exhibits sparser but larger spikes, which tend to reach or even, surpass the RMSE of the worst ranking model (i.e., the MLP). The KAN and the MLP are inferior. 

\paragraph{Summary of main lessons.} The common denominator in all of the above benchmarks is the superior performance of the 1D-TFNO. It is consistently the top, or among the top performers, it exhibits stable RMSE on different test sizes, it can learn efficiently high-frequency features (including noise) and has consistent performance across very different situations without the need for excessive hyperparameter tuning. We have observed similar qualitative features in a series of other functions as well, and believe this provides a clear motivation towards a more systematic use of NOs as function approximators/interpolators, especially in higher-dimensional settings where the training data are necessarily sparse. In a nutshell, learning functions across higher-dimensional subspaces appears to be superior to point-wise learning.

\section{Application: Nuclear Binding Energies}
\label{sec:nuclear}

\begin{figure*}[t]
    \centering
    \includegraphics[width=\textwidth]{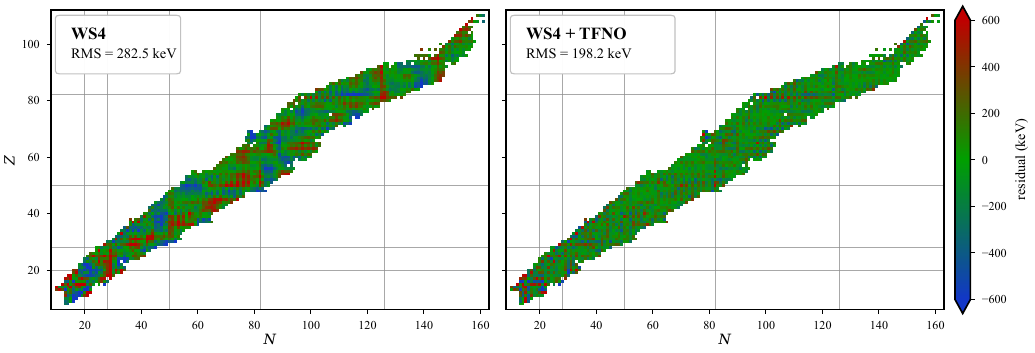}
    \caption{Nuclear-chart residual fields for the strict AME2020--WS4 subset. Left: the raw WS4 residual, $E_b^{\mathrm{exp}}-E_b^{\mathrm{WS4}}$. Right: the held-out residual after the 30-member neural-operator (NO) correction, $E_b^{\mathrm{exp}}-(E_b^{\mathrm{WS4}}+\Delta E_b^{\mathrm{NO}})$. Horizontal and vertical guide lines mark proton and neutron magic numbers.}
    \label{fig:nuclear}
\end{figure*}

Atomic nuclei are characterized by their proton and neutron numbers, $(Z,N)$, and their binding energy $E_b(Z,N)$ is one of the central observables of nuclear physics. 
It governs nuclear stability, separation thresholds, and decay $Q$-values. It is also a key input to broader physics applications, including astrophysical reaction-network calculations for the rapid-neutron-capture ($r$-)process, which synthesized roughly half of the heavy elements in the universe~\citep{Burbidge:1957vc}. 
These calculations probe regions of the nuclear chart where no direct measurements exist, and for many applications uncertainties of a few hundred keV already matter~\citep{Burbidge:1957vc,PhysRevLett.116.121101,Mumpower:2015ova}.

Over the past decades, nuclear mass modeling has advanced through several complementary frameworks: from the original liquid-drop~\citep{Weizsacker:1935bkz}, and nuclear shell models~\citep{GoeppertMayer1949} to modern global macroscopic descriptions refined by microscopic corrections~\citep{Goriely:2013xba,Wang:2014qqa,Moller:2015fba}.
These models capture the global structure of the mass surface remarkably well, but their residual errors still exhibit coherent organization across the $(Z,N)$ chart.
In other words, after subtraction of a strong baseline, the remaining problem is not an arbitrary high-dimensional regression task but a structured residual-field interpolation problem. This observation has motivated a growing body of machine-learning corrections to physics mass models~\citep{Niu:2018csp,Niu:2022gwo,Wu:2022nnc,Gao2021Correction,Le:2023siy,Lu:2024nkr,Huang:2025rfg,Bentley:2024bnm,Bentley:2025fbj,Agrawal:2025rvn,Kitouni:2023rct}. 

Moreover, recent interpretability studies~\citep{Kitouni:2023rct,Kitouni:2024ulw,Richardson:2025dze}, sharpen two lessons. 
First, high-capacity models recover familiar local nuclear structure from data, which suggests that the residual mass surface is locally smooth and highly structured on the chart. 
Second, quoted RMS values are only meaningful when each target nucleus is evaluated out of sample, with no leakage of its binding energy into training. Motivated by these points, in this work we train TFNOs to learn the residuals to one of the state-of-the-art nuclear mass models, namely the latest Weizs{\"a}cker-Skyrme model (WS4) and evaluate only with strictly out-of-sample predictions, following the unbiased-evaluation principle emphasized by ~\citet{Richardson:2025dze}.

\paragraph{Dataset.}
We use experimental binding energies from the Atomic Mass Evaluation 2020 (AME2020)~\citep{Huang:2021nwk}, matched nucleus by nucleus to predictions from the Weizs\"acker--Skyrme version-4 (WS4) mass model~\citep{Wang:2014qqa}. We train on the subset of AME2020 nuclei satisfying the chart cut $N>12$ or $Z>12$. For evaluation, we further restrict to the well-measured region relevant for sub-MeV mass modeling by requiring an experimental uncertainty below $100\,\mathrm{keV}$. This leaves $2{,}348$ nuclei in the testable subset.

Rather than predict $E_b$ directly, we learn the WS4 residual
\begin{equation}
    \Delta E_b(Z,N) = E_b^{\mathrm{exp}}(Z,N) - E_b^{\mathrm{WS4}}(Z,N),
\end{equation}
and report the root-mean-square (RMS) error of the corrected prediction $E_b^{\mathrm{WS4}}+\Delta E_b^{\mathrm{TFNO}}$ against experiment. On this subset, raw WS4 has a pooled five-fold RMS of $282.47\,$keV.

\paragraph{Chart completion.}
We place the data on the $(Z,N)$ grid, with proton number on one axis and neutron number on the other. The grid has shape $(111,162)$, but only the known nuclei are included in the support mask. Each pixel carries five input channels: normalized $Z$, normalized $N$, the visible WS4 residual, a visible-data mask, and a support mask indicating where a real nucleus exists. Training is masked inpainting on the chart. For each outer fold, validation nuclei are never visible. Within the remaining training nuclei, each batch randomly hides about $5\%$ of the known residuals, shows the other $95\%$, and computes the loss only on the hidden residuals. At evaluation time the held-out fold is predicted from the chart coordinates, support, and the residuals of the training nuclei. Thus the model is not asked to memorize single points; it is asked to complete a partially observed physical field.

\paragraph{Architecture and metric.}
Each TFNO member uses $n_{\mathrm{modes}}=(24,32)$, $h=96$ hidden channels, $L=2$ spectral layers, Tucker factorization with rank $r=0.01$, five input channels, and one output channel, for $146{,}929$ trainable parameters. The 30-member ensemble has $4.4$ million trainable parameters in total, while each member remains small enough to train in minutes on a single graphics processing unit (GPU). We train with AdamW using learning rate $3\times 10^{-4}$, weight decay $10^{-4}$, batch size $4$, $64$ steps per epoch, and early stopping over at most $800$ epochs.

We report pooled OOF RMS over five folds. Each nucleus is predicted exactly once by an ensemble for which that nucleus was held out, and the fold predictions are pooled into one RMS over the full strict subset. This follows the logic of the unbiased metric emphasized in~\citet{Richardson:2025dze}: in-sample RMS can be arbitrarily optimistic for high-capacity ML models, while a single train/test split can depend strongly on which hard nuclei happen to land in the test set. In our protocol, the held-out residuals are never placed in the visible-context channel.

\paragraph{Results and comparison with the literature.}
A single TFNO member already reaches $208.3\pm2.7\,$keV pooled five-fold OOF RMS (mean $\pm$ std over $46$ independent seeds). The model is also cheap to train: best checkpoints occur within the first few tens of epochs, and each member is a minutes-scale single-GPU training job. Because the seeds are independent, ensembling adds almost no engineering complexity: the $30$-member TFNO ensemble, trained embarrassingly in parallel, reaches $198.2\,$keV---a $30\%$ reduction relative to WS4. The raw WS4 residual field and the held-out residuals after the neural-operator correction are shown in \cref{fig:nuclear}. 

Direct comparison across the nuclear-mass ML literature is delicate because the quoted RMS depends on the underlying mass table, subset cuts, train/test protocol, and the information made available to the model during training. We therefore organize \cref{fig:nuclear_comparison} into two blocks. The top block collects \emph{coordinate-only single-task} models and serves as the fairest comparator set for our setup. Within this block, TFNO+WS4 outperforms most reported architectures, with the notable exception of NuCLR~\citep{Kitouni:2023rct} in its single-task realization.\footnote{We note that \citet{Richardson:2025dze} argue that AI effectively rediscovered a physics-inspired local interpolation principle---which they term \emph{Jaffe factorization}---and trace it back to the classic work of \citet{Garvey:1966zz}. Remarkably, when implemented as local corrections to WS4 on the same evaluation set, this principle yields state-of-the-art interpretable performance, reaching an RMS below $120\,\mathrm{keV}$, comparable to or better than full neural-network models.} The bottom block collects models that are either \emph{multi-task} (MT) representation learners, \emph{physics-informed} (PI) systems using engineered pairing, shell, or deformation features, or richer ensemble/conflation schemes that combine residual learners across multiple baseline mass models~\citep{Agrawal:2025rvn}. We quote published RMS values as reported in the original references, preferring clearly held-out or \emph{chronological} evaluations\footnote{This means training on an older Atomic Mass Evaluation release and testing on nuclei first appearing in a newer release, e.g.\ AME2012$\rightarrow$AME2020. Such protocols use a much smaller effective test frontier: from AME2012 to AME2020 the newly added nuclei are only about $5.8\%$ of AME2020, far below the $\sim 20\%$ test fraction of standard splits.} when available.

\begin{figure}[t]
    \centering
    \scriptsize
    \setlength{\tabcolsep}{1pt}
    \begin{tabular}{@{}>{\raggedright\arraybackslash}p{0.45\linewidth}@{\hspace{0.3em}}r@{\hspace{0.1em}}l r@{}}
        \hline
        Method, evaluation protocol & \multicolumn{1}{r}{\hspace{1em}RMS [keV]} & & size \\
        \hline
        \multicolumn{4}{@{}l}{\textit{Physics baseline}} \\
        WS4, this subset~\citep{Wang:2014qqa} & 282.5 & \rmsbar{gray}{1.45cm} & --- \\
        \hline
        \multicolumn{4}{@{}l}{\textit{NNs, coordinate-only single-task (fair comparators)}} \\
        FNN+LDM, held-out~\citep{Le:2023siy} & 370 & \rmsbar{gray}{1.90cm} & $1551$ \\
        BNN+WS4, held-out~\citep{Niu:2018csp} & 212 & \rmsbar{gray}{1.09cm} & $169$ \\
        CNN+WS4, chrono.~\citep{Lu:2024nkr} & 211 & \rmsbar{gray}{1.08cm} & not reported \\
        \textbf{TFNO+WS4, 5-fold OOF} & \textbf{198.2} & \rmsbar{blue}{1.01cm} & \textbf{4.4M} \\
        NuCLR, 100-fold OOF~\citep{Kitouni:2023rct,Richardson:2025dze} & 130 & \rmsbar{gray}{0.67cm} & $10$M \\
        \hline
        \multicolumn{4}{@{}l}{\textit{Other ML techniques or PI / MT - trained NNs}} \\
        LSBET+multi-model+PI, indep. AME2020 test~\citep{Bentley:2024bnm} & 199 & \rmsbar{orange}{1.02cm} & $3$k trees \\
        LightGBM+WS4+PI, held-out~\citep{Gao2021Correction} & 170 & \rmsbar{orange}{0.87cm} & $50$k trees \\
        FMTE+multi-model+PI, indep. AME2020 test~\citep{Bentley:2025fbj} & 164 & \rmsbar{orange}{0.84cm} & $12$k trees \\
        KRR+WS4+MT, LOFOCV/extrap.~\citep{Wu:2022nnc} & $150$ & \rmsbar{orange}{0.77cm} & $6$k \\
        FCNN+PI, held-out, $Z,N{>}20$~\citep{Huang:2025rfg} & 122 & \rmsbar{orange}{0.63cm} & $4.4$--$9$k \\
        NuCLR+MT, 100-fold CV~\citep{Kitouni:2023rct} & 79 & \rmsbar{orange}{0.41cm} & $10$M \\
        \hline
    \end{tabular}
    \caption{Published RMS values (keV) for recent nuclear-mass ML models under held-out, chronological, pooled OOF, or reported cross-validation protocols, shown relative to the WS4 baseline on our strict subset. The top block is the fair comparator set for this work: coordinate-only single-task models on WS4- or LDM-residual targets, evaluated under explicit held-out, chronological, or pooled OOF protocols. The bottom block gathers models that are either multi-task (MT, sharing information across other observables, \emph{e.g.} separation energies, radii etc., at the same $(Z,N)$), physics-informed (PI, with engineered pairing, shell, or deformation features or residuals to multiple physics mass models), or richer ensemble schemes, and are shown for context only. Blue denotes this work. Subset cuts differ slightly between rows. Where possible, the final column reports audited parameter counts; for tree-based models it instead reports learner counts or ensemble composition.}
    \label{fig:nuclear_comparison}
\end{figure}

\section{Discussion and Conclusions}
\label{sec:conclusions}

We have introduced a novel framework that recasts neural operators---originally designed for operator learning in infinite-dimensional function spaces---as efficient interpolators of finite-dimensional functions. The key insight is a simple auxiliary-base-space construction that lifts any function into an operator mapping amenable to NO architectures.

In our benchmarks on analytic functions, we compared the zero-dimensional and one-dimensional TFNOs against MLPs and KANs on examples of target functions with various degrees of complexity and dimensionality.

As a real-world demonstration, we applied our framework to nuclear binding energies---a problem of fundamental importance in nuclear and astroparticle physics. Casting the nuclear chart as a partially observed 2D residual field, a 30-member TFNO ensemble reaches state-of-the-art performance in pooled five-fold OOF evaluation. This demonstrates that NOs can extract meaningful physical corrections while keeping the evaluation protocol explicitly held out. More generally, this example highlights a practical advantage of the NO-as-interpolator viewpoint: the single models are small, train quickly, and scale to ensembles without multi-GPU synchronization or hand-engineered physics features.

Future directions include: (i) combining the NO interpolator with multi-task learning for simultaneous prediction of multiple nuclear observables; (ii) testing prospective extrapolation to newly measured or dripline nuclei rather than only interpolation within the known chart; and (iii) extending the theoretical analysis to establish formal approximation bounds for NOs in the function-interpolation setting.

\section*{Acknowledgments}
We are grateful to James Chryssanthacopoulos, Alex Stapleton and Mike Williams for useful discussions and comments on the manuscript. ST is supported by the Swiss National Science Foundation project number P5R5PT\_222350, and acknowledges CERN TH Department for hospitality while this research was being carried out. This work is also supported by the National Science Foundation under Cooperative Agreement PHY-2019786 (The NSF AI Institute for Artificial Intelligence and Fundamental Interactions, http://iaifi.org/).

\bibliography{workshop_refs}
\bibliographystyle{icml2026}

\newpage
\appendix

\section{Additional Benchmark Plots}
\label{app:plots}

In this appendix, we collect additional plots from the benchmarks presented in Sec.~\ref{sec:benchmarks}.

\begin{figure}[h!]
    \centering
    \includegraphics[width=\linewidth]{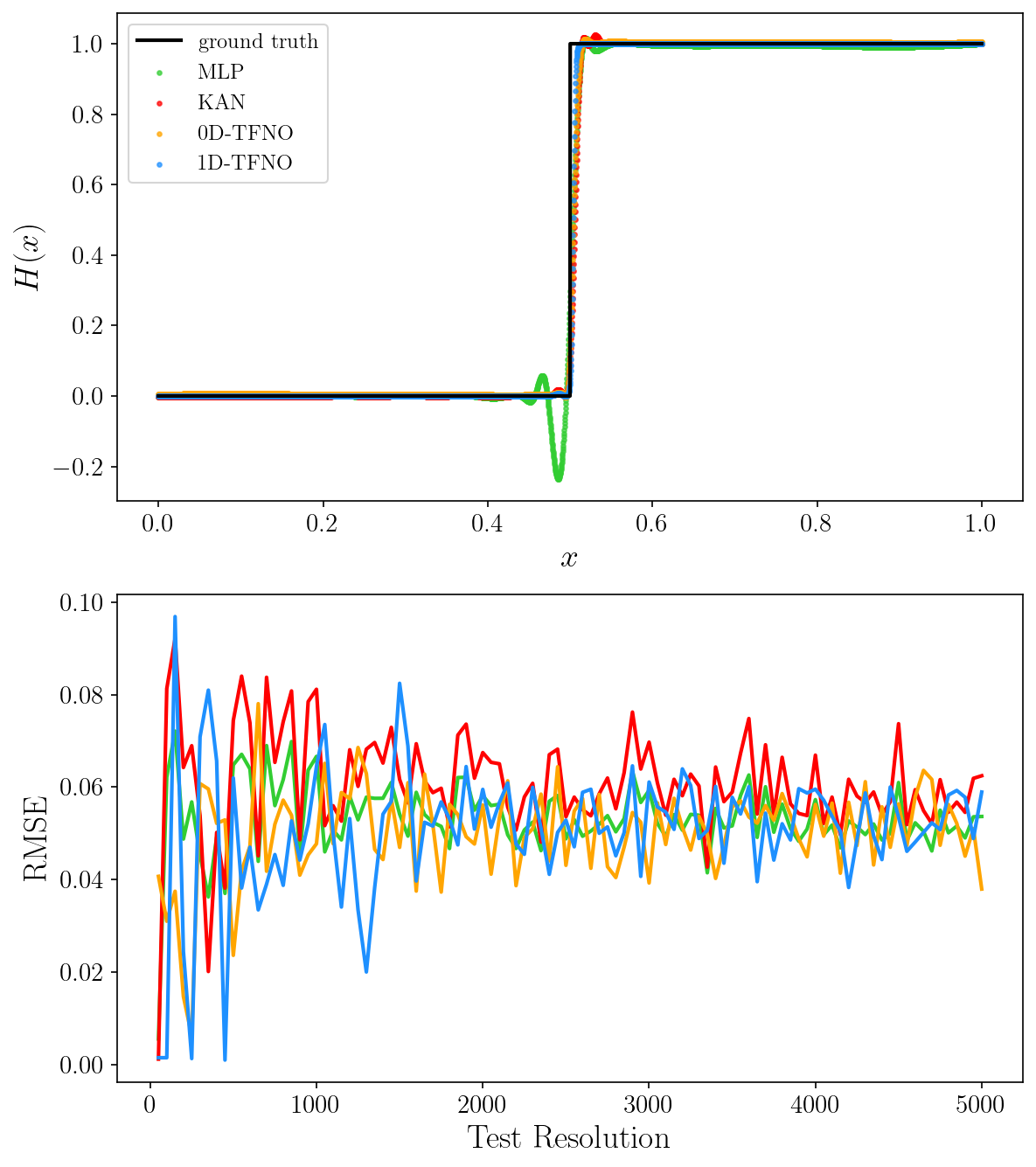}
    \caption{Benchmark results for the Heaviside function. Top plot: Predicted maps of all models against the function argument, $x$. Bottom plot: test-set RMSE vs number of test points.
    }
    \label{fig: step benchmarks}
\end{figure}

\begin{figure*}[h]
    \includegraphics[width=\linewidth]{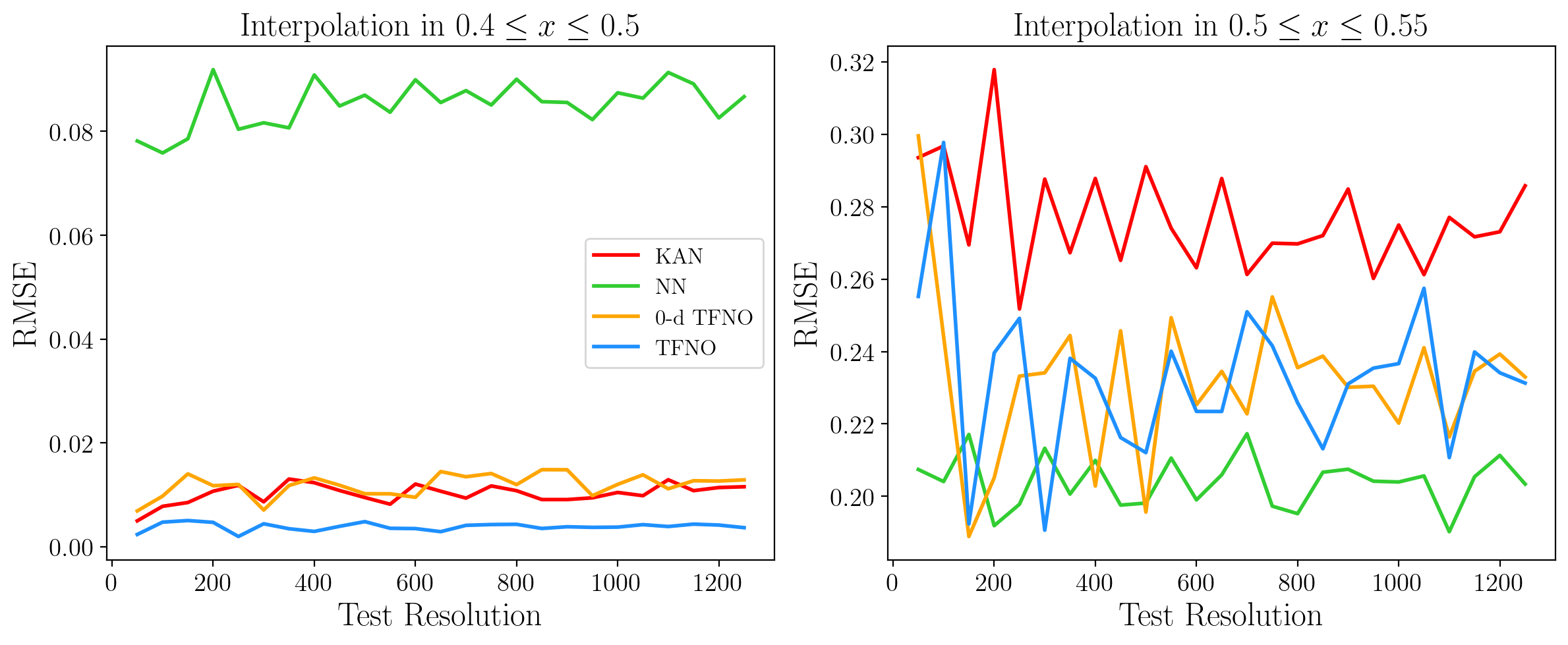}
    \caption{Zoomed test-set RMSE near the Heaviside discontinuity, shown separately in the intervals $0.4 \le x \le 0.5$ and $0.5 \le x \le 0.6$.}
    \label{fig: heaviside rmses}
\end{figure*}

\begin{figure}[h!]
    \centering
    \includegraphics[width=\linewidth]{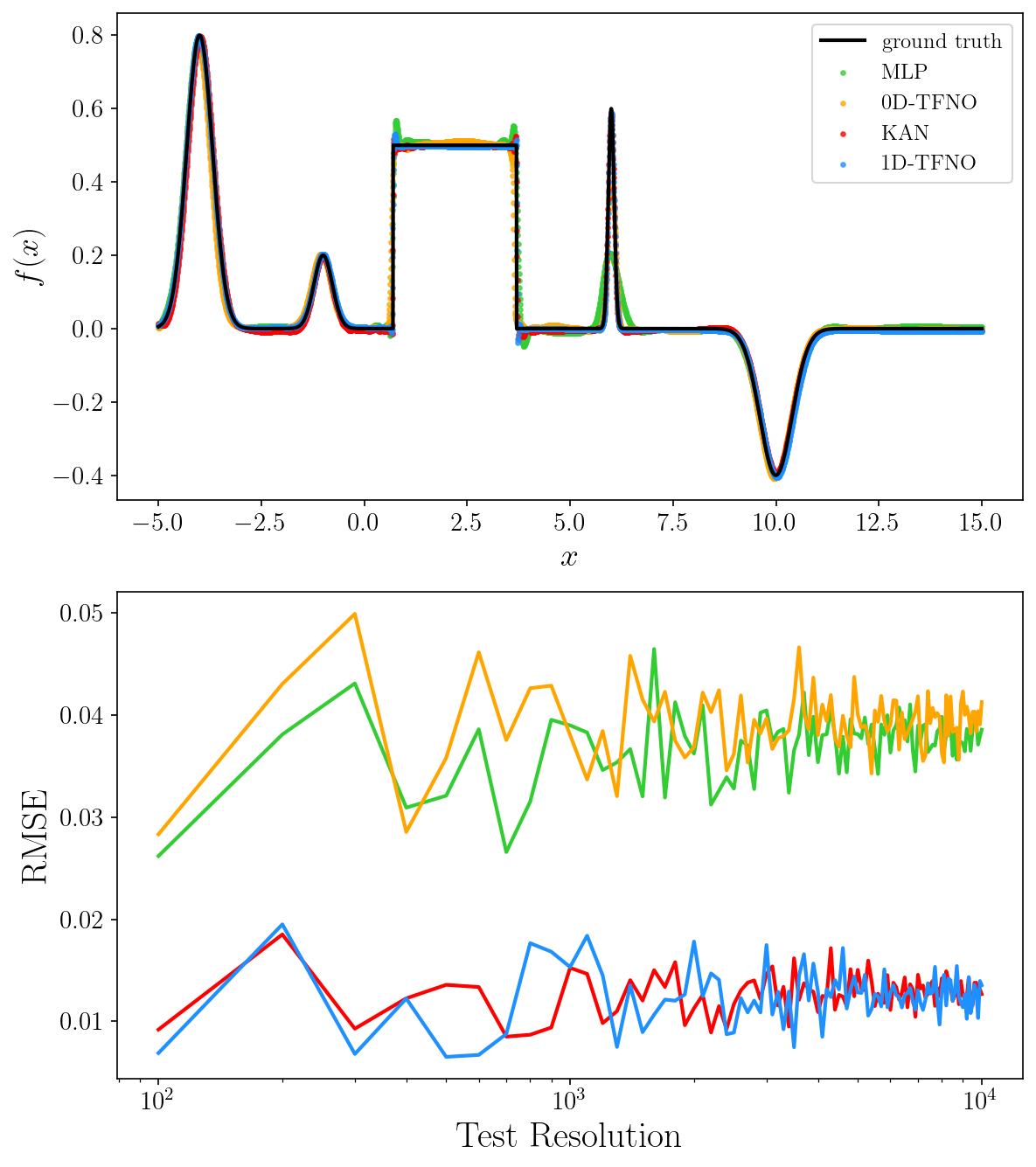}
    \caption{Benchmark results for the piecewise composition. Top plot: Predicted maps of all models against the function argument, $x$. Bottom plot: test-set RMSE vs number of test points. 
    }
    \label{fig: piecew comb}
\end{figure}

\begin{figure}[h!]
    \centering
    \includegraphics[width=1\linewidth]{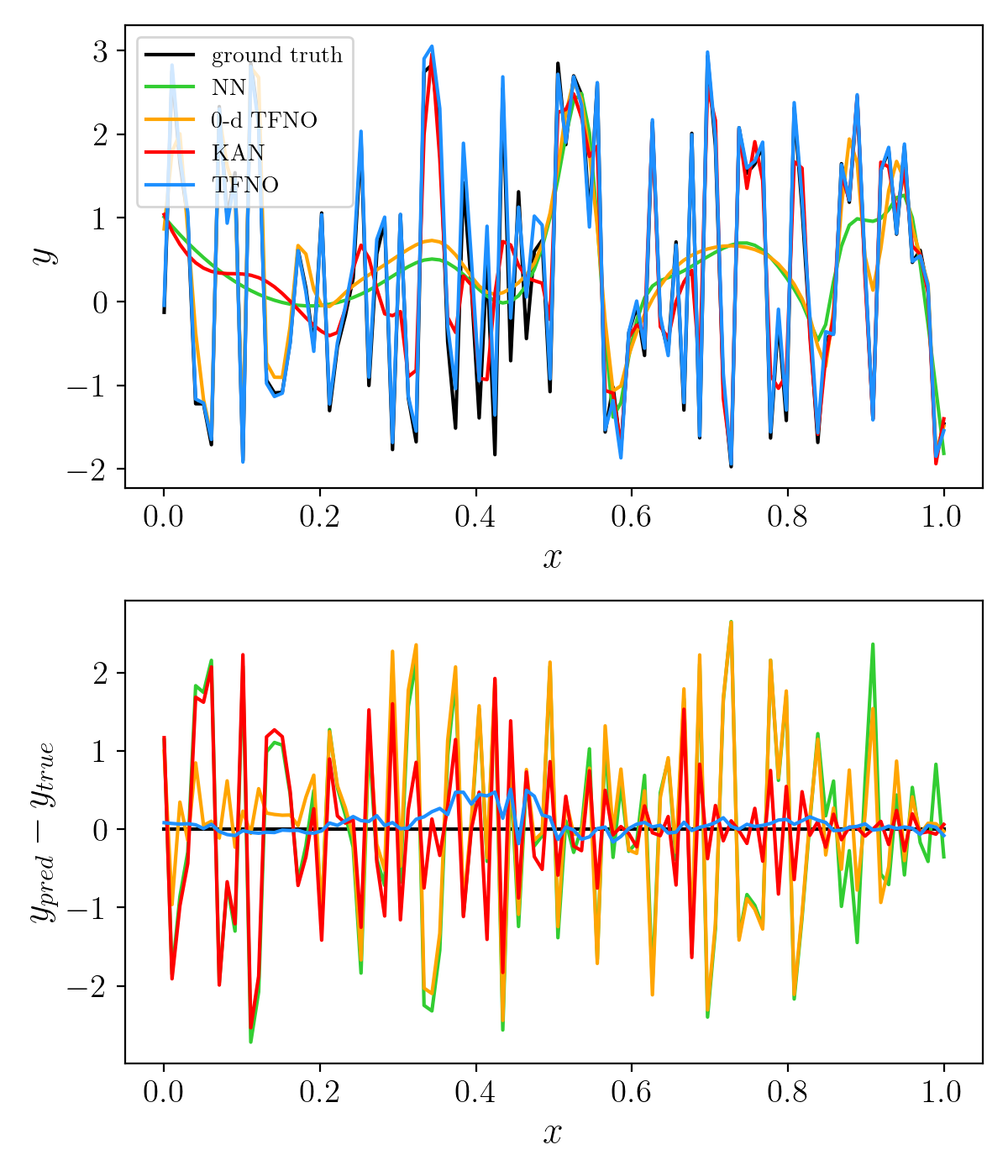}
    \caption{Benchmark results for 1D uniform noise. Top plot: Predicted maps of all models vs. the function argument, $x$. Bottom plot: residuals of all models vs. $x$.
    }
    \label{fig: noise}
\end{figure}

Figs.~\ref{fig: step benchmarks}, \ref{fig: heaviside rmses} contain the results obtained on the fits of the Heaviside function. Fig.~\ref{fig: piecew comb} contains the fits of a target function that combines Gaussian and step functions and Fig.~\ref{fig: noise} refers to the analysis of uniform noise. All examples in this appendix are one-dimensional functions.

\end{document}